\documentclass{hotnets22}

\usepackage{cite}
\usepackage{times}  
\usepackage{hyperref}
\usepackage{makecell}

\hypersetup{pdfstartview=FitH,pdfpagelayout=SinglePage}

\usepackage[english]{babel}

\usepackage[compact]{titlesec}
\titlespacing{\subsection}{0pt}{0ex}{0ex}
\titlespacing{\subsubsection}{0pt}{0ex}{0ex}
\titlespacing{\section}{0pt}{0ex}{0ex}

\usepackage[linesnumbered,ruled,vlined]{algorithm2e}
\usepackage[noend]{algpseudocode}
\usepackage{array}
\newcolumntype{x}[1]{>{\centering\arraybackslash\hspace{0pt}}p{#1}}
\usepackage{amsmath}
\usepackage[inline]{enumitem}
\usepackage{epsfig}
\usepackage{fancyvrb}
\usepackage{graphicx}
\usepackage{booktabs, tabularx}
\usepackage[flushleft]{threeparttable}
\usepackage{latexsym, color}
\usepackage{listings}
\usepackage{tikz}
\usepackage{url}
\usepackage[font=normalsize,labelfont=bf]{caption}
\usepackage{hyperref}
\usepackage{cleveref}
\usepackage{tabularx}
\usepackage{subcaption}
\usepackage{xcolor}
\usepackage{multirow}
\usepackage{multicol}
\usepackage{float}
\usepackage{adjustbox}

\crefformat{figure}{Figure~#2#1#3}
\definecolor{commentgreen}{RGB}{2,112,10}
\definecolor{eminence}{RGB}{108,48,130}
\definecolor{weborange}{RGB}{255,165,0}
\definecolor{frenchplum}{RGB}{129,20,83}

\long\def\comment#1{}

\usepackage{cuted}
\usepackage{float}
\usepackage{xspace}
\newcommand{\ie}{{\em i.e.}}
\newcommand{\eg}{{\em e.g.}}

\newcommand{\para}[1]{\noindent {\bf #1}}
\newcommand{\qiao}[1]{\textcolor{red}{[qiao: #1]}}

\usepackage{array}
\usepackage{setspace}

\begin{document}

\title{
Toward Reproducing Network Research Results \\
Using Large Language Models 
}

\author{\large
 Qiao Xiang$^{\ast}$,
 Yuling Lin$^{\ast}$, 
 Mingjun Fang$^{\ast}$,
 Bang Huang$^{\ast}$,
 Siyong Huang$^{\ast}$,\\
 \large
 Ridi Wen$^{\ast}$, 
 Franck Le$^{\dagger}$, 
 Linghe Kong$^{\diamondsuit}$,
 Jiwu Shu$^{\ast\circ}$\\
 \large
 $^\ast$ Xiamen University,
 $^\dagger$IBM Research,
\\
\large
$^{\diamondsuit}$ Shanghai Jiao Tong University,
$^{\circ}$Minjiang University
\\
}

\maketitle

\begin{abstract}
Reproducing research results in the networking community is important for both
academia and industry. The current best practice typically resorts to three
	approaches: (1) looking for publicly available prototypes; (2)
	contacting the authors to get a private prototype; and (3) manually
	implementing a prototype following the description of the publication.
However, most published network research does not have public
prototypes and private prototypes are hard to get. As such, most reproducing
	efforts are spent on manual implementation based on the publications,
	which is both time and labor consuming and error-prone.
	In this paper, we boldly propose reproducing network research
	results using the emerging large language models (LLMs). 
	In particular, we
first prove its feasibility with a small-scale experiment, in which
four students with essential networking knowledge each
reproduces a different networking system published in prominent conferences and
journals by prompt engineering ChatGPT. We report the experiment's observations and lessons and discuss future open research questions 
of this proposal. This work raises no ethical issue.
\end{abstract}


\section{Introduction}\label{sec:intro}

Reproducing network research results have both significant education and
research values. For education, it completes students'
learning process on computer networks with lecture attendance and textbook
reading, in accordance with the usual process of science study
worldwide, \eg, educators at Stanford University assign reproduction projects
in their networking classes~\cite{yan2017learning}. For research, 
it ensures the results are accurate and trustworthy, gives researchers a
hands-on opportunity to understand the pros and cons of these results, and
motivates more innovations, \eg, IMC 2023 introduces a replicability track for
submissions that aims to reproduce or replicate results previously
published at IMC~\cite{imc23}.

The best practice for people to reproduce a published network research typically involves one of three approaches. First, rerun a publicly available prototype provided by
 the authors (\eg,~\cite{ap-ton16}) or other people who implement it based on the publication (\eg,~\cite{apkeep-nsdi20}). Second, if
no public prototype is available, people may contact the authors to ask for a private prototype. Third, if no prototype is
available, people need to manually implement one following the publication.

\para{The best practice to reproduce network research results has limitations.}
All three approaches in the best practice are limited for various reasons.
First, not much published research comes with a publicly available prototype.
Our study shows that even in prominent networking conferences such as SIGCOMM and NSDI,
only a small number of papers provide publicly available prototypes from the authors. From 2013 to 2022, only 32\% and 29\% of papers in SIGCOMM and NSDI, respectively, provide open-source prototypes.
Although some non-authors implement prototypes and release them to the public~\cite{flashcode}, the number of such prototypes is even smaller. 
Second, the authors sometimes are reluctant to share a private prototype 
for various reasons (\eg, patent filing, commercial product, policy, and security). 

As such, without ready-made prototypes, the dominant way for people to reproduce the
results of a published networking paper is to manually implement its proposed
design. Although this "getting-hands-dirty" approach provides people precious
experience in understanding the details of the paper, in particular the pros and
cons of the proposed design, the whole process is both time and labor consuming
and error-prone. In the long run, it is unsustainable because network research
results are becoming more and more complex. If people are spending more time
trying to reproduce the published results, they will have  less time for
critical thinking and innovation. One may think this is not a unique issue for the networking
community, but a prevalent one for the whole computer science discipline.
However, the situation is more severe for networking research. For example, it
may take a fresh graduate student one week to reproduce a machine learning paper
by manual implementation, but one or two months to reproduce a networking paper.

\para{Proposal: reproducing network research results using large language
models (LLMs).} 
In this position paper, we make a bold proposal to reproduce network research
results by prompt engineering the emerging LLMs. Such a proposal, if
built successfully, can benefit the networking community from multiple
perspectives, including (1) substantially simplifying the reproduction process,
(2) efficiently identifying missing details and potential vulnerabilities (\eg,
hyper-\\parameters and corner-case errors) in network research results, and (3) 
motivating innovations to improve published research. It could even help improve
the peering review process of networking conferences~\cite{shenker2022rethinking}, partially realizing the vision of a SIGCOMM April Fools' Day email in 2016~\cite{sigcomm-joke-2016}.

Our proposal is backed by the recent success of applying LLMs to both
general~\cite{zhang2022repairing, zhang2022automated, verbruggen2021semantic,
10.1145/3485535, nijkamp2022conversational, copilot} and domain-specific code
intelligence tasks~\cite{yen2021semi, chen2022software, le2022rethinking} as
evidences. For general programming, Copilot~\cite{copilot} can
provide effective code completion suggestions.  ChatGPT~\cite{chatgpt} can complete and debug simple coding tasks when given
proper prompts.  Rahmani et al. ~\cite{10.1145/3485535} integrate LLM and
content-based synthesis to enable multi-modal program synthesis. For domain-specific programming, in particular the network domain, \\SAGE
~\cite{yen2021semi}
uses the logical form of natural-language sentences to identify ambiguities in RFCs and automatically generate RFC-compliant code.
NAssim~\cite{chen2022software} uses an LLM to parse network device manuals
and generate corresponding configurations for devices. They both focus on well-formatted inputs with a limited range of topics (\ie, RFC and manuals).

In this paper,
we go beyond to take a first step to thoroughly investigate the feasibility and challenges of
reproducing network research results by prompt engineering LLMs.




\para{A preliminary experiment~(\S\ref{sec:exp}).} 
We conduct a small-scale experiment, where we choose four networking systems
published in prominent networking conferences and journals~\cite{ncflow,
arrow-te, ap-ton16, apkeep-nsdi20} and ask four students with essential knowledge of
networking to each reproduce one system by prompt engineering the free
ChatGPT~\cite{chatgpt}, a publicly available chatbot built on GPT-3.5, a
representative LLM~\cite{gpt3.5}. 

The results verify the \textit{feasibility} of our proposal, 
\ie, each student successfully reproduces the
system assigned to her / him via ChatGPT. Their correctness is validated by comparing the results of small-scale test cases with those of the corresponding open-source prototype. The efficiency
is evaluated using large-scale datasets. Results show that their efficiency is similar to that of their open-source prototypes.

We learn several lessons from the experiment on how to use LLMs to
reproduce networking research results more efficiently. First, provide LLMs with separate, modular prompts to build different components of a
system and then put them together, rather than provide monolithic prompts to build the whole system. Second, ask LLMs to implement components with
pseudocode first to avoid unnecessary data type and structure
changes later. Third, data preprocessing is important to reproduction, but is
often missed in the paper. We also learn some guidelines for
debugging LLM-generated code, including sending error messages /
error test cases to LLMs, and specifying the correct logic in more detailed
prompts.

\para{Open research questions (\S\ref{sec:challenges}).}
We identify several key open research questions regarding LLM-assisted network research results reproduction and elaborate on opportunities to tackle them. They include: (1) how to handle the diversity of these results  (2) how to design a (semi-) automatic prompt engineering framework to reproduce
these results?  (3) how to identify and handle the missing details and vulnerabilities of these results?  (4) 
how to develop a domain-specific LLM for network research reproduction?
(5) how to use LLMs to discover optimization opportunities for
these results?  (6) how to apply this approach of reproduction to promote
computer networking education and research?

\begin{figure}[t] 
    \centering
\includegraphics[width=0.46 
 \textwidth]{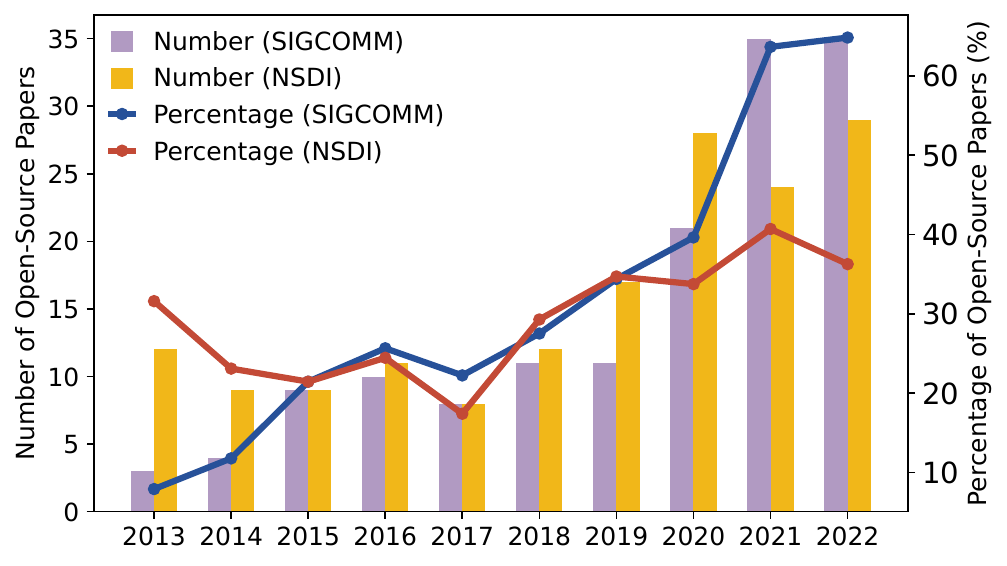} 
    \caption{Statistics of SIGCOMM and NSDI papers with an open-source prototype from the authors (2013 - 2022).}
\label{fig.conference}
    \end{figure}

\section{Background and Motivation}\label{sec:background}
We conduct a study on network research reproduction in two prominent
network conferences (\S\ref{sec:reproduce}) and
elaborate on the motivation of our proposal with a simple example
(\S\ref{sec:motivation}).

\subsection{Background}\label{sec:reproduce}
\para{A statistical study on network research results reproduction.} 
We collect the full research papers in SIGCOMM and NSDI of the past 10 years
(\ie, 2013 - 2022). For each paper, we 
collect: (1) whether the proposed
system is open-sourced by the authors; (2) how many other systems are compared in the evaluation;
(3) how many of them are open-source prototypes? 
\footnote{We do not differentiate publicly available and private prototypes
because it is hard to collect this information.}; 
and (4) how many of them are reproduced by the authors of the proposed system.
Figure~\ref{fig.conference} plots the open-source statistics of SIGCOMM and NSDI.
In total, only 32\%/29\%/31\% of SIGCOMM/NSDI/both papers
in the past 10 years release open-source prototypes. 

\begin{figure}[t] 
    \centering
\includegraphics[width=0.48\textwidth]{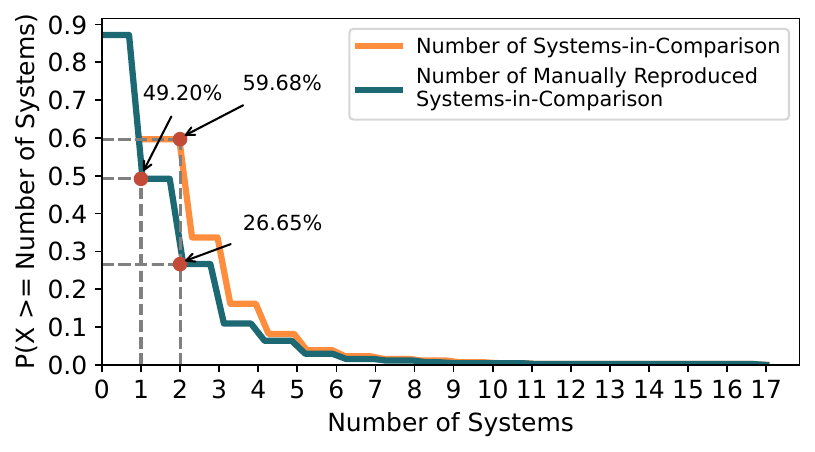} 
    \caption{Statistics of the number of compared systems and manually reproduced ones of each paper.}
\label{fig.cdf}
    \end{figure}

Figure~\ref{fig.cdf} plots the statistics of the number of systems-in-comparisons of each paper and how many of them require manual implementation. 
We observe that the authors spend substantial efforts to manually implement the systems of others. 59.68\% of papers compare with at least two other systems.
On average, the authors of each paper have to manually reproduce 2.29 systems. 49.20\% / 26.65\%  of the papers have to manually reproduce at least one / two other systems.

\para{Existing reproduction approaches are insufficient.} 
Our study indicates that \textit{although there is a universal need
in the community to reproduce network research results, the most dominant way to
do so is still manual implementation by following the publication.}
Although it would provide people with precious
hands-on experiences to better understand the published network research
results, it is time and labor consuming and error-prone. It is
unsustainable because network research results, in particular the system ones, are becoming more and more complex. The more time people 
spend on network research results reproduction, the less time they are left with for critical thinking and innovation.

\subsection{Reproducing Using LLMs: A Motivating Example}\label{sec:motivation}
Inspired by the recent success of LLMs in code intelligence tasks (\eg, code
completion~\cite{copilot},  debug~\cite{chatgpt, zhang2022repairing,
zhang2022automated}, and synthesis~\cite{10.1145/3485535, yen2021semi}), we
propose to prompt engineering the emerging LLMs to reproduce network research
results. As a motivating example,
we let one undergraduate senior majoring in computer science interact with free ChatGPT.
to develop a UDP server and client that interact to play rock-paper-scissors. After only four prompts with a total of 159 words, ChatGPT generates
the correct program for both the server and client, with a total of 93 lines of code (LoC) in Python. Figure ~\ref{fig.socket} shows two snippets of the generated program. Despite a simple example, it shows
that an LLM can implement network programs and gives us
the confidence to experiment with more complex network research results.

\definecolor{dkgreen}{rgb}{0,0.6,0}
\definecolor{gray}{rgb}{0.5,0.5,0.5}
\definecolor{mauve}{rgb}{0.58,0,0.82}

\lstdefinestyle{Python}{
    frame=tb,
    language=Python,
    aboveskip=3mm,
    belowskip=3mm,
    showstringspaces=false,
    columns=fixed,
    basicstyle={\scriptsize\ttfamily},
    numbers=none,
    numberstyle=\tiny\color{gray},
    keywordstyle=\color{blue},
    commentstyle=\color{dkgreen},
    stringstyle=\color{mauve},
    breaklines=true,
    breakatwhitespace=true,
    tabsize=3
}


\begin{figure}[t]
\centering
\begin{lstlisting}[style=Python]
def run_server():
  host = '127.0.0.1'
  port = 12345

  server_socket = socket.socket(socket.AF_INET, socket.SOCK_STREAM)
  server_socket.bind((host, port))
  server_socket.listen(1)

  print("Server is running...")

  round_number = 1

  while True:
    client_socket, addr = server_socket.accept()
    print("Connected to", addr)

    while True:
      client_message = client_socket.recv(1024).decode('utf-8')
      ...

def run_client():
  host = '127.0.0.1'
  port = 12345

  client_socket = socket.socket(socket.AF_INET, socket.SOCK_STREAM)
  client_socket.connect((host, port))

  print("Connected to the server.")

  while True:
    guess = input("Enter your guess (P/R/S for paper/rock/scissors, or D to disconnect): ")
    guess = validate_input(guess)
    ...
\end{lstlisting}
\caption{Code snippets of a UDP server and client playing rock-paper-scissors generated by ChatGPT.}
\label{fig.socket}
\end{figure}


\section{A Preliminary Experiment}\label{sec:exp}
To investigate the feasibility 
of LLM-assisted network research results reproduction, we conduct a small-scale experiment participated by four
students. 

\subsection{Methodology}
\para{Participants with basic computer science knowledge.} 
The four participants (referred to as $A$, $B$, $C$, and $D$) are students from
three research universities in China. $A$ is a first-year master's student majoring in computer science, with a focus on interpretable machine
learning.  The other three are senior undergraduates who will start their master's
program in computer science in September 2023. During their undergraduate
study, $A$, $B$, and $C$ major in computer science, and $D$ majors in
information and computing science. Before the experiment starts, they all have received basic training in programming, operating
systems, and computer networks and have basic English communication skills.

\begin{figure}[t] 
    \centering
\includegraphics[width=0.46\textwidth]{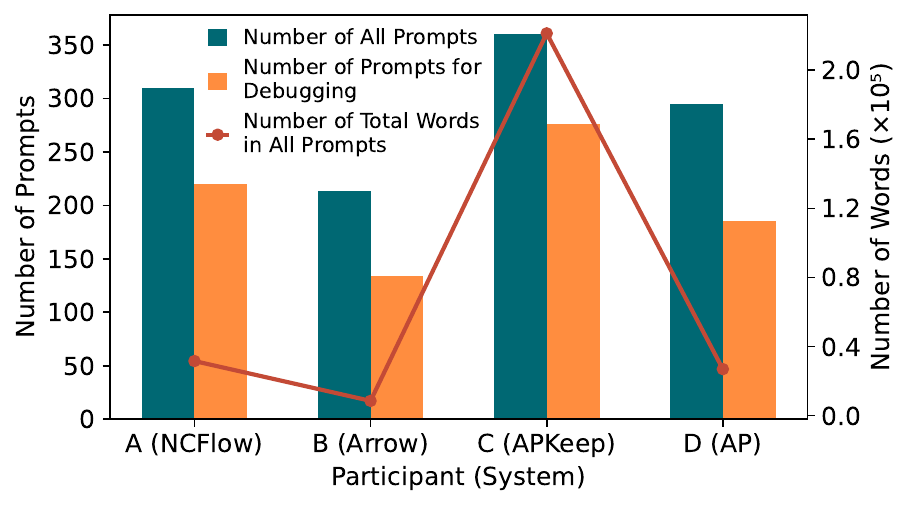} 
    \vspace{-0.5em}
    \caption{The number of prompts used by four participants.}
\label{fig.prompt}
    \end{figure}


         


\para{Deciding which systems to reproduce.}
We focus on four systems, two traffic engineering (TE)
systems (NCFlow~\cite{ncflow} and Arrow~\cite{arrow-te}) and two data plane verification (DPV)
systems (AP~\cite{ap-ton16} and APKeep~\cite{apkeep-nsdi20}). All of them are published in top-tier
conferences and journals (\ie, SIGCOMM, NSDI, and ToN). We choose them for three reasons. First, they are all software systems that can be reproduced in
general-purpose programming languages (\eg, Java and Python). Existing LLMs may
understand and generate programs in these languages
better than those in domain-specific programming languages (\eg, P4 and Verilog). Second,
they all run in a centralized controller, which has a simpler architecture and
is easier for LLMs to understand than distributed systems (\eg, BGP and Paxos).
Third, the papers describing them provide clearly structured
details of the systems (\ie, modular components, workflow, definitions,
formulations, and pseudocode), making it easier for participants to understand the systems and design proper prompts to interact with LLMs.

\para{Experiment procedure.}
We mimic the procedure of the reproduction project in CS244 of
Stanford University~\cite{yan2017learning}. Each participant is assigned one
different aforementioned system and asked to reproduce it using
free ChatGPT
in a 25-day period
during spare time. Participants can choose Java or Python as the language
for reproducing. They are not allowed to write the implementation code themselves or
manually modify the code generated by ChatGPT, nor to copy the prompts of
others. However, they can 
write their own test cases and are encouraged to discuss with
each other and other students about the details of the papers and how to interact with ChatGPT. We meet with them online every three to five days to discuss their progress. During these meetings, we do not answer any questions about
what specific prompts the participants should use. Instead, we provide them with
suggestions on how to reproduce the system from a system designer's perspective.
Participants are also allowed to ask us questions about the papers, but they do not do so. 

\para{Validating the reproduction.}
Participants validate their reproduction
by comparing it with the open-source prototypes of these systems. Three
systems have open-source prototypes from their authors~\cite{apcode,ncflowcode,arrowcode}  and the fourth has a non-author open-source prototype 
~\cite{flashcode}
whose correctness and efficiency are validated in~\cite{flash}. Each participant uses small-scale test cases to examine the correctness of their reproduction. If the output is inconsistent with that of the open-source prototype, she / he manually 
analyzes the root cause 
and interacts with ChatGPT to fix the errors. Afterward,
the participant evaluates
its performance on large-scale datasets and compares it with that of the open-source prototype.

\subsection{Results}\label{sec:results}
\begin{figure}[t] 
    \centering
\includegraphics[width=0.45\textwidth]{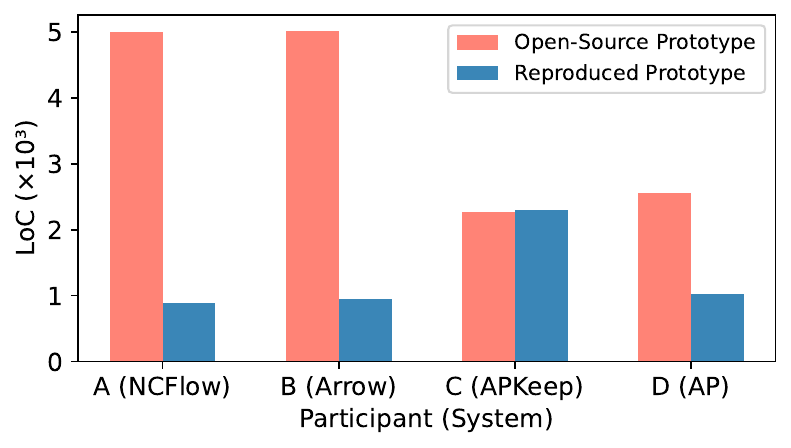} 
\vspace{-0.5em}
    \caption{The LoC of open-source /  reproduced prototypes.}
\label{fig.code}
    \end{figure}

\para{Reproducing network research results via LLMs is feasible.}
All four participants successfully reproduce their assigned systems. Their log is in \cite{log}. The correctness is validated by comparing them with the open-source prototypes in small-scale test cases. Figure~\ref{fig.prompt} shows the number of prompts and words each participant used during reproducing. Figure~\ref{fig.code} compares the LoC of the reproduced and open-source prototypes. We next
present findings on their performance. 

\para{Participant $A$: reproducing NCFlow with high accuracy and performance.}
$A$ evaluates the performance of reproduced NCFlow in 13 TE instances from ~\cite{ncflow}. The reproduced NCFlow computes the  objective function value with a
maximal of 3.51\% difference from the open-source prototype and a maximal
end-to-end computation latency of 6.4 seconds. Although the latency is up to
111x higher, it is due to the choice of LP solvers between two implementations
(\ie, the reproduced one uses Pulp~\cite{pulp} and the open-source one uses
Gurobi~\cite{gurobi}). The LoC of the reproduced prototype is only 17\% of that of
the open-source one. It is because the open-source prototype has a large portion
of code for formatting and parsing the irregular inputs of datasets.

\para{Participant $B$: reproducing Arrow with low accuracy due to paper-code inconsistency.} 
$B$ evaluates the reproduced Arrow in two TE instances from~\cite{arrow-te}. The computed objective function has an up to 30\% difference from that of the open-source prototype.  After comparing the source code of the two prototypes, we find that the
root cause is the inconsistency between the paper description and the
open-source implementation. For example, some predefined parameters in the paper
are implemented as decision variables in the open-source prototype. The
definition of the restorable tunnel in the paper is also different from the open-source prototype. $B$ interacts with ChatGPT to
reproduce Arrow based on the paper, leading to the performance discrepancy.
The reproduced Arrow has a much smaller LoC (19\%) because (1) $B$ uses Python, which has a richer library than Julia used by the open-source prototype and (2) when receiving prompts, ChatGPT tends to generate shorter code.  

\para{Participant $C$: reproducing APKeep accurately with comparable latency.} 
$C$ compares the reproduced APKeep with a non-author open-source prototype on
four real-topology datasets to verify loop-free, blackhole-free reachability. In
all cases, both prototypes compute the same number of atomic predicates and have
approximately the same latency. Both  use JDD~\cite{jdd} for
binary decision diagram operations and have approximately the same number of LoC. 

\para{Participant $D$: reproducing AP accurately but with worse performance due
to the choice of different BDD libraries and missing details in the paper.}
$D$ compares the reproduced AP with the open-source one on
three real-topology datasets to verify loop-free, blackhole-free reachability.
Although both compute the same number of atomic predicates and
verification results, the reproduced AP has a substantially worse latency: (1) up to 20x longer time to compute predicates and (2) up to $10^4$x longer time to verify
reachability). The root cause of the former is the use of JavaBDD, a library with a worse performance of BDD operations than JDD. The root cause of the latter is the missing details of the 
reachability verification algorithm in the paper. The paper only gives the algorithm on given a path, how to find the predicates
reaching $d$ from $s$. It does not describe how to efficiently find all
the predicates reaching $d$ from $s$ from any path (\eg, the authors use a selective BFS traversal in their open-source prototype). 
Because $D$ is not a computer science major and unfamiliar with the exponential complexity
of path enumeration, $D$ decides to use the algorithm in the paper as a building block to enumerate all the paths from $s$ to $d$ and check the
reachable predicates for each path, leading to a much higher verification latency. This could be avoided by stating the use of
selective BFS traversal in the paper.
We later discuss how we may integrate LLM-assisted reproduction and formal methods to  identify such missing details.  
\definecolor{dkgreen}{rgb}{0,0.6,0}
\definecolor{gray}{rgb}{0.5,0.5,0.5}
\definecolor{mauve}{rgb}{0.58,0,0.82}
\lstdefinestyle{Java}{
    frame=tb,
    language=Java,
    aboveskip=3mm,
    belowskip=3mm,
    showstringspaces=false,
    columns=fixed,
    basicstyle={\tiny\ttfamily},
    numbers=none,
    numberstyle=\tiny\color{gray},
    keywordstyle=\color{blue},
    commentstyle=\color{dkgreen},
    stringstyle=\color{mauve},
    breaklines=true,
    breakatwhitespace=true,
    tabsize=3
}



\begin{figure*}[]
\setlength{\abovecaptionskip}{0cm}
\setlength{\belowcaptionskip}{-0.cm}
\begin{subfigure}{0.36\linewidth}
\setlength{\abovecaptionskip}{0.18cm}
\setlength{\belowcaptionskip}{-0.cm}
	    \centering\includegraphics[width=1\linewidth]{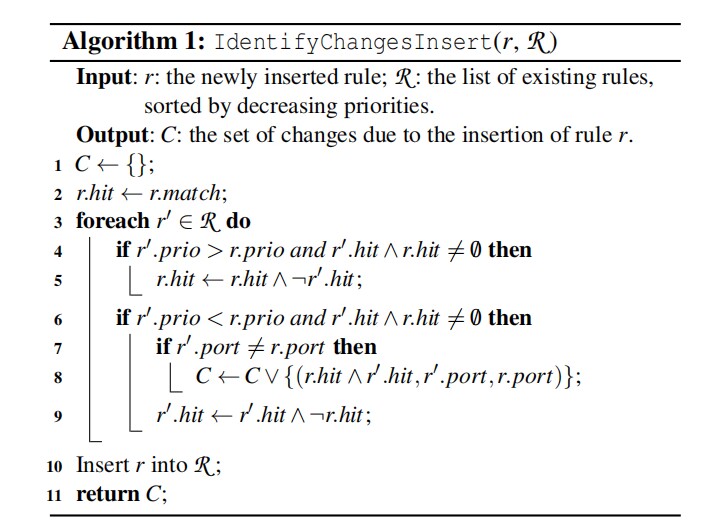}
		\caption{\label{fig:pseudo-code} Pseudocode in \cite{apkeep-nsdi20}.}
	\end{subfigure}
\begin{subfigure}{0.4\linewidth}
\setlength{\abovecaptionskip}{0.cm}
\setlength{\belowcaptionskip}{-0cm}
\begin{lstlisting}[style=java]
if (r.getPriority() > rule.getPriority()) {
    int newHit = bddEngine.diff 
    (rule.getHit(), r.getHit());
    bddEngine.deRef(rule.getHit());
    rule.setHit(newHit);
}
if (r.getPriority() < rule.getPriority()) {
    int intersection = bddEngine.and 
    (r.getHit(), rule.getHit());
    int newHit = bddEngine.diff (r.getHit(), intersection);
    bddEngine.deRef(r.getHit());
    r.setHit(newHit);
    if (intersection != BDDEngine.BDDFalse && r.getOutPort() != rule.getOutPort()) {
        changes.add(new Change 
        (intersection,r.getOutPort(),rule.getOutPort()));
    } else {
        bddEngine.deRef(intersection);
    }
}
\end{lstlisting}
\caption{\label{fig:open-source-apkeep} Manually implemented code ~\cite{apcode}.}
\end{subfigure}
\begin{subfigure}{0.23\linewidth}
\setlength{\abovecaptionskip}{0cm}
\setlength{\belowcaptionskip}{-0cm}
\begin{lstlisting}[style=java]
if (r.getPrio() > rule.getPrio()) {
    rule.setHit(bddEngine.diff
    (rule.getHit(), r.getHit()));
    }
    
int inter = bddEngine.and(r.getHit(), rule.getHit());
if (r.getPrio() < rule.getPrio()) {
    r.setHit(bddEngine.diff
    (r.getHit(), inter));
    if (inter != BDDEngine.BDDFalse 
    && r.getPort() != rule.getPort()) {
        changes.add(new Change
        (inter, r.getPort(), rule.getPort()));} else {
        bddEngine.deRef(inter);
        }
    }
\end{lstlisting}
		\caption{\label{fig:chatgpt-apkeep} ChatGPT-generated code.}
	\end{subfigure}
\caption{The pseudocode, manually implemented code and ChatGPT-implemented code of APKeep to identify all behavior changes caused by a rule insertion.}
	\label{fig:alg-and-code}
\vspace{-2em}
\end{figure*}



\subsection{Lessons}
After the experiment, we interview the participants to gather their experience
and summarize several lessons.

\para{Asking LLMs to implement a system component by component, not the whole system all at once.} 
In the beginning, all participants tried to send ChatGPT
prompts like "implement XX that works in the following steps XXX".
ChatGPT does not respond well to such \textit{monolithic} prompts. As such, they
switch to a top-down approach, which divides the system into 
components, and for each component, sends ChatGPT more detailed \textit{modular} prompts 
in sequence to implement, debug and test it. This allows them to reproduce the system successfully. It shows that ChatGPT's capability of
understanding system design and implementing it is still limited to small systems
and components.

\para{Implementing components with pseudocode first.} 
Given a paper, the part that is the closest to the real code (and sometimes
the authors' thought) is its pseudocode (Figure~\ref{fig:alg-and-code}).  As such, it would be ideal if ChatGPT
could receive it as a prompt to generate the code using it as a basis. However,
suppose we first ask ChatGPT to implement the components without pseudocode.
When asking ChatGPT to implement the components with pseudocode later,
the generated code often requires substantial changes to data types and structures in the previously generated ones. It is because ChatGPT implements components described by text-based prompts differently from the ones described by pseudocode-based prompts, leading to interoperability issues.
Two participants find that implementing components with pseudocode first allows ChatGPT to stabilize the key data
types and structures and avoid changing them when implementing other
components. As such, 
we plan to design
\textit{a pseudocode-like intermediate
representation for all components of a system} to improve the efficiency of
LLM-assisted reproduction.

\para{Data preprocessing is important to the system, but not to the research paper.} 
Because information on data preprocessing is usually not provided by the
research paper, participants have to learn about it by looking into the datasets
and send ChatGPT data format-related prompts based on their understanding.
As such, a generic and automatic data preprocessing solution is another key for
improving the efficiency of LLM-assisted reproduction.

\para{Three guidelines for debugging.}
First, participants find that many bugs they encounter 
can be fixed by sending the error messages of compiler / runtime to ChatGPT. Many
 such bugs are data type errors and can be  avoided by
explicitly specifying key variables' data types and structures 
or at least describing their needed operations and properties 
in prompts. Second, for simple
logic bugs, sending the test case causing them to ChatGPT can be an effective
approach to repair them. Third, for more complex logic bugs, we can repair them
by specifying the correct logic of the code in more detailed, sometimes
step-by-step prompts.

\section{Open Research Questions}\label{sec:challenges}
We discuss some open research questions of our proposal.

\para{Handling the diversity of published network research.}
This diversity comes from two folds. First, network research papers have very diverse organization, content, and level of detail.
For example, SIGCOMM and NSDI papers are usually very system-heavy, while the ones in other venues
(\eg, TON, ICNP and INFOCOM) focus more on the analytical side. Reproducing papers from different venues may require different
ways of processing and digesting them. Second,  networking is an area
spanning many topics, including not only  layers of the network
architecture stack but also programmable hardware and distributed systems.
Reproducing papers on different topics may require different ways of prompt
engineering. Previous studies 
focus on a limited range of topics and use well-formatted input (\ie, RFCs~\cite{yen2021semi} and manuals~\cite{chen2022software}). As such, their design may not
apply to the broader context of reproducing network research results.

One direction to tackle this issue is a unified prompt
engineering framework for reproducing network research results~\cite{luca2022prompting,jules2023chatgpt}. One design is to
follow the top-down approach of system development with the following steps:
(1) describe to the LLM the key components of the system, (2) describe
how components interact and ask the LLM to define the 
interfaces, (3) provide the LLM with the details of each component to generate the code, (4) test and debug the LLM-implemented component, (5) repeat (3) and (4) for
each component, and (6) test and debug the complete system.

\begin{sloppypar}
\para{Improving the efficiency of reproducing via (semi-) automatic prompt
engineering LLMs.}
Although the human-LLM interaction in manual prompt engineering is beneficial for people to better
understand the details of published research results, the efficiency of
reproducing can be substantially improved by (semi-) automatic prompt
engineering~\cite{yongchao2022large}. 
Our investigation on this
issue focuses on automating the aforementioned unified prompt engineering
framework. In particular, given a  network research paper, our initial
design involves (1) using an LLM for natural language (\eg, ChatPDF~\cite{chatpdf}) to
understand the paper and extract its architectures, key components, and workflow,
(2) transforming the extracted information into multi-modal prompts (\eg, logic
predicates, pseudocode, examples, and test cases) for the overall architecture
and each component, (3) sending the prompts for components to an LLM for coding
(\eg, FlashGPT3 ~\cite{verbruggen2021semantic}) to generate, test and debug codes
for each component, and (4) sending the prompts for architecture to piece different components into a complete system. 

\end{sloppypar}

\para{Handling missing details and vulnerabilities in published network research.}
Due to the space and time limit, the authors may have to omit non-essential technical
details in their published research. Such missing details pose additional
challenges for reproducing these results \eg,  the experience of participants $B$ and $D$ in our experiment (\S\ref{sec:results}). 

Published network research may also have
vulnerabilities, such as inaccurate descriptions of designs and examples (\eg,
\cite{flash}) and hard-to-tune hyper-parameters (\eg, \cite{apkeep-nsdi20}). They 
make it challenging for LLMs to reproduce the results correctly (\eg, sending an incorrect example to LLMs).

We may resort to formal
methods (\eg, \cite{bibel2013automated, rival2020introduction}) in two ways to identify such missing details and vulnerabilities. First, we could verify and analyze the workflow and algorithms extracted from the paper by either human
efforts or the LLMs to search for such issues~\cite{zhang2023performal, yen2021semi}. Second, if a paper has an open-source prototype, we may
comparatively analyze it and the LLM-reproduced one to examine their
functionality and performance discrepancy using modular
checking~\cite{tang2021campion}.

\para{Building domain-specific LLMs for network research reproduction.}
Such LLMs can substantially improve the scope and efficiency of LLM-assisted
network research reproduction. In our  experiment, participants reproduce centralized software systems using ChatGPT, a chatbot built on
a general-purpose LLM. However, many network research results propose hardware
systems~\cite{sivaraman2016programmable, li2016clicknp}, hardware-software
co-design systems~\cite{netcache, univmon} and distributed systems~\cite{raft,
zookeeper}. Not only are they more complex to reproduce, some of them
also require domain specific languages (\eg, P4 and Verilog) that are very different from general programming languages (\eg, Java and Python). As
such, general-purpose LLMs may have difficulty reproducing these systems.

We propose to specifically build a network research reproduction LLM by
using network research materials (\eg, papers, codes, and RFCs on various network
research topics) as training data. Early evidence supporting the feasibility of
such an LLM is the recent success of programming-oriented LLMs
~\cite{zhang2022repairing, zhang2022automated, verbruggen2021semantic,
10.1145/3485535, nijkamp2022conversational, copilot, yen2021semi,
chen2022software, le2022rethinking} in providing code completion suggestions,
identifying and fixing bugs, and reproducing RFCs. One key challenge is the availability of data. Although there are huge amounts of
network traffic data~\cite{le2022rethinking}, 
there is substantially less network research code available. One way to tackle this is to 
integrate data augmentation and static analysis to produce more network research code.


\para{Discovering innovation opportunities from reproducing networking research
results.}
LLM-assisted networking research reproduction could help reveal innovation
opportunities. First, this process could deepen researchers' understanding on
published research results, helping organize their intellectual and critical
thinking.  Second, analyzing the reproduced prototype using automatic program
analysis~\cite{zhang2023performal, scalpel-apnet23} could expose bottlenecks of
the proposed design, leading to system optimization opportunities. Third,
although it may still be a long shot, it is theoretically feasible to build a deep
learning model with open-source and reproduced prototypes of networking
research as datasets to predict networking innovations, similar to the recent
building of AlphaFold~\cite{jumper2021highly} in the area of biology. To this
end, interpretable machine learning on networking systems would be an important
tool~\cite{meng2020interpreting, uint} given their capability to extract logic
rules and explanations behind system behaviors.

\para{Promoting computer networking education and research.}
In the era where AI is the predominant computer science research area, 
reproducing network research results using LLMs could draw students' attention and motivate their interest in networking education and research. First, by interacting with LLMs 
they get not only a hands-on experience with the latest AI
breakthrough, but also the opportunity to understand the
classic and latest network research results. As such, it is useful
for students to strengthen their career (both academia and industry) skills in
both computer networks and AI. Second, as discussed
earlier, this process could help students discover innovation opportunities
in networking research. Third, it fits into the competency-building model of the recent computer science education
paradigm~\cite{clear2017cc2020, cs2023}.
Last but not least, LLM-assisted  network research results reproduction could help improve the peering review process of prominent networking
conferences~\cite{shenker2022rethinking}. One may recall an
April Fools' Day email in the SIGCOMM mailing list in 2016 saying that SIGCOMM
will introduce AI to automate the paper review process.
Although we believe that this email is still too "forward-looking", 
our proposal would be an interesting trial to automate part of
the review process. 

LLM-assisted network research reproduction, in particular a
(semi-)automatic framework, also has a negative impact: students may misuse it to finish projects.
How to walk the fine line between leveraging LLMs-assisted reproduction to
promote networking education and research and misusing it is an 
important question for both academia and industry.

\section{Conclusion}\label{sec:conclusion}
We propose to simplify network
research results reproduction using LLMs. We validate its feasibility with a small-scale experiment, where students successfully reproduce
networking systems 
by prompting engineering
ChatGPT. We summarize our lessons 
and discuss open
research questions.



\newpage
\bibliographystyle{abbrv}
\begin{small}

\end{small}
\end{document}